\documentclass[runningheads]{llncs}
\usepackage[T1]{fontenc}
\usepackage{graphicx}
\usepackage{algorithm}
\usepackage{algorithmic}
\usepackage{kotex}
\usepackage{multirow}
\usepackage{booktabs}
\usepackage{array}
\usepackage{makecell}
\usepackage{multicol}
\usepackage{nicefrac}  
\usepackage{microtype} 
\usepackage{mathtools}
\usepackage{amsfonts}
\usepackage{dsfont}
\usepackage{cuted}
\usepackage{capt-of}
\usepackage{amsmath,amssymb}
\usepackage{subfig} 
\usepackage{xspace}
\usepackage{cases}
\usepackage{pifont}
\usepackage[dvipsnames]{xcolor}
\usepackage{hyperref}

\usepackage{color}

\newcommand{\cmark}{\ding{51}}%
\newcommand{\xmark}{\ding{55}}%
\DeclareMathOperator{\E}{\mathbb{E}}
\newcommand{\ourdata}{{Cephalometric X-ray}\xspace}
\newcommand{\spinewebdata}{AASCE\xspace}

\newcommand{\ourmethod}{{Ours}\xspace}
\newcommand{\oursemethod}{Interaction-guided gating\xspace}

\makeatletter
\newcommand{\printfnsymbol}[1]{%
  \textsuperscript{\@fnsymbol{#1}}%
}

\begin{document}
\title{Morphology-Aware \\Interactive Keypoint Estimation}
\titlerunning{Morphology-Aware Interactive Keypoint Estimation}
\author{
Jinhee Kim\inst{1}\thanks{Both authors contributed equally.} \and 
Taesung Kim\inst{1}\printfnsymbol{1} \and 
Taewoo Kim\inst{1} \and 
Jaegul Choo\inst{1}\and \\ 
Dong-Wook Kim\inst{2} \and 
Byungduk Ahn\inst{3} \and 
In-Seok Song\inst{2}\thanks{Both authors are the co-corresponding authors.} \and 
Yoon-Ji Kim\inst{4}\printfnsymbol{2} 
}

\authorrunning{J. Kim et al.} 
\institute{
KAIST, Daejeon, South Korea \\
 \email{ \{seharanul17, zkm1989, specia1ktu, jchoo\}@kaist.ac.kr}\\ 
\and
Korea University Anam Hospital, Seoul, South Korea \\ 
\email{densis@korea.ac.kr}
\and
Papa's dental clinic, Seoul, South Korea\\
\and
Asan Medical Center, Ulsan University School of Medicine, Seoul, South Korea\\
\email{yn0331@gmail.com}
}

\maketitle              
\setcounter{footnote}{3}
\renewcommand{\thefootnote}{\fnsymbol{footnote}}

\begin{abstract}
Diagnosis based on medical images, such as X-ray images, often involves manual annotation of anatomical keypoints. However, this process involves significant human efforts and can thus be a bottleneck in the diagnostic process.
To fully automate this procedure, deep-learning-based methods have been widely proposed and have achieved high performance in detecting keypoints in medical images. 
However, these methods still have clinical limitations: accuracy cannot be guaranteed for all cases, and it is necessary for doctors to double-check all predictions of models. In response, we propose a novel deep neural network that, given an X-ray image, automatically detects and refines the anatomical keypoints through a user-interactive system in which doctors can fix mispredicted keypoints with fewer clicks than needed during manual revision.
Using our own collected data and the publicly available \spinewebdata dataset, we demonstrate the effectiveness of the proposed method in reducing the annotation costs via extensive quantitative and qualitative results. A demo video of our approach is available on our \href{https://seharanul17.github.io/interactive_keypoint_estimation/}{\textbf{project webpage}}\footnote{\url{https://seharanul17.github.io/interactive_keypoint_estimation/}}.
\end{abstract}

\section{Introduction}

\begin{figure}[t]
\begin{center}
    \includegraphics[width=1.0\linewidth]{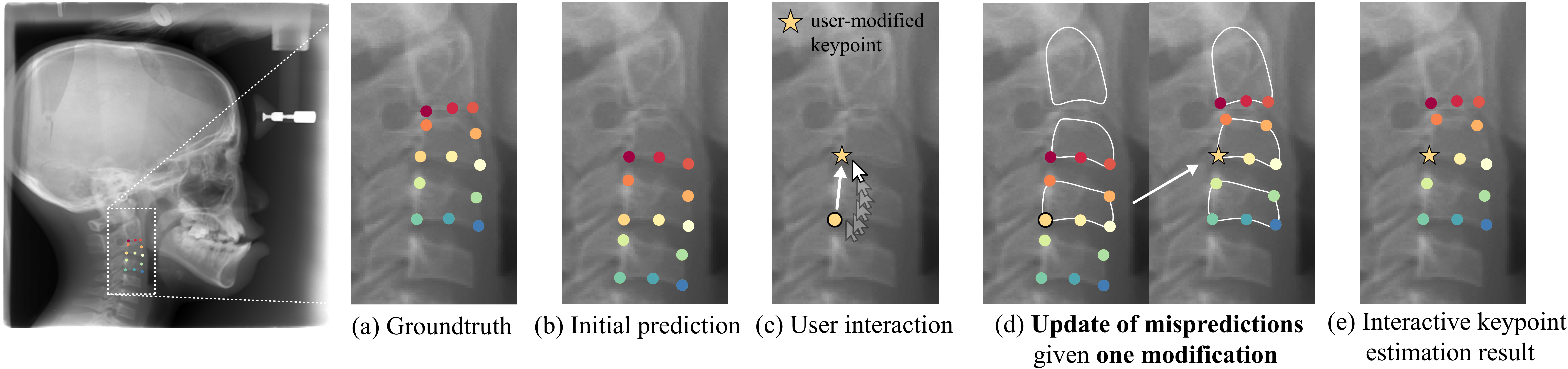}
\end{center}
  \caption{{Interactive keypoint estimation results} on a cephalometric X-ray image.
The goal is to estimate (a) 13 keypoints on the cervical vertebrae, each of which determines vertebrae morphology.
Here, (b) the initial prediction misses one vertebra at the top, making the entire prediction wrong.
A manual revision will be no better than annotating from scratch.
However, in our method, if (c) a user corrects only one point, (d) the remaining points come up together. All keypoints appropriately reflect user-interaction information in (e) the final result.
}
\label{fig:overview}
\end{figure}

Keypoint-based analysis of medical images is a widely used approach in clinical treatment~\cite{bier2018x,kim2021prediction,kordon2019multi,med_1,spinewebdataset}. 
Anatomical points obtained from an image can be utilized to measure significant features of body parts where keypoints are located.
For example, vertex points of cervical vertebrae can be used to measure the concavity and height of the vertebrae, which correlates to skeletal growth~\cite{kim2021prediction,med_1}. 
Usually, doctors manually annotate such keypoints, but the process is costly, requiring significant human effort. 
Accordingly, studies on fully automatic keypoint estimation approaches have been widely conducted~\cite{recent3,topology,recent2,cephann,isbi,spinewebdataset,recent4}.
However, fully automatic methods can be inaccurate because data available for training is usually scarce in the medical domain due to privacy issues.  
Besides, errors of inaccurate keypoints can also cascade to subsequent feature extraction and diagnosis procedures.
Thus, a thorough revision of erroneous model predictions is crucial, yet manually correcting individual errors is highly time-consuming. 
To assist the process, a keypoint estimation model that, after receiving user correction for a few mispredicted keypoints, automatically revises the remaining mispredictions is needed.
However, to the best of our knowledge, little research has been conducted on such an approach despite its need and importance. 
In response, we propose a novel framework, the \textbf{interactive keypoint estimation method}, which aims to reduce the number of required user modifications compared to manual revision, thereby reducing human labor in the revision process, as shown in Fig.~\ref{fig:overview}.  

Recently, interactive segmentation tasks have gained much attention for their ability to reduce human effort in precise image segmentation of medical image datasets~\cite{inter2,inter1,2021miccai_interactiveSeg1} as well as real-world datasets~\cite{brs,firstclick,fbrs}.
Reviving iterative training with mask guidance (RITM)~\cite{ritm} achieves the state-of-the-art performance in image segmentation by reactivating iterative training for multiple user revisions. 
Also, approaches that optimize user revision during the inference time, such as BRS~\cite{brs} and f-BRS~\cite{fbrs}, have shown remarkable performance in the interactive segmentation task.
Given these successes, we develop our interactive keypoint estimation model from the interactive segmentation approaches.

However, when we apply these methods to the interactive keypoint estimation task, we find that information of user modifications to keypoints does not propagate to distant points in the model; it only locally affects nearby keypoints.
To address this issue, we propose the \textbf{interaction-guided gating network}, which can propagate user modification information across the spatial dimensions of the image.
The proposed gating network better reflects user modification information in all mispredictions, including distant keypoints. 
We also observe that the degree of freedom between keypoints is minimal in medical images, e.g., vertex points on the cervical vertebrae have limited deformation. In Fig.~\ref{fig:overview}, the cervical vertebrae consist of a series of similarly shaped polygonal vertebrae with five corner points.
This indicates that the distances or the angles between keypoints may be similar across different images. 
If so, we can use these higher-order statistics obtained from keypoints to make the model explicitly aware of dependent relationships between keypoints.
To this end, we propose \textbf{the morphology-aware loss}, which facilitates explicit learning of inter-keypoint relationships by regularizing distances and angles between predicted keypoints to be close to the groundtruth.

Consequently, our main contributions are as follows: (\lowercase\expandafter{\romannumeral1}) We propose a novel interactive keypoint estimation network in which users can revise inaccurate keypoints with fewer modifications than needed when manually revising all mispredictions. 
(\lowercase\expandafter{\romannumeral2}) We introduce an interaction-guided gating network, which allows user modification information to be propagated to all inaccurate keypoints, including distant ones.
(\lowercase\expandafter{\romannumeral3}) We propose the morphology-aware loss, which utilizes higher-order statistics to make the model aware of inter-keypoint relationships as well as individual groundtruth locations of keypoints.
(\lowercase\expandafter{\romannumeral4}) We verify the effectiveness of our approach through extensive experiments on a dataset we collected on our own as well as on the public \spinewebdata challenge dataset~\cite{spinewebdataset}.

\begin{figure*}[t!]
\begin{center}
  \includegraphics[width=1.0\linewidth]{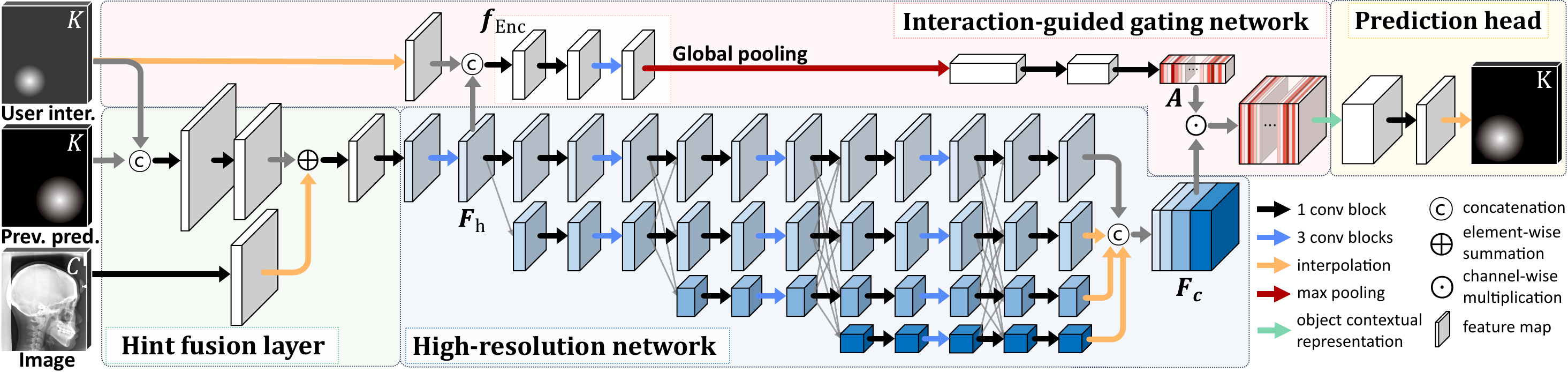}
\end{center}
  \caption{{Overview of proposed interactive keypoint estimation model.} 
  It receives an image, user interaction (User inter.), and its previous prediction (Prev. pred.) and outputs a heatmap of keypoint locations that reflects interactive user feedback.
}
\label{fig:main_network}
\end{figure*}

\section{Methodology}\label{method}

\subsection{Interactive Keypoint Estimation}\label{key}
 Interactive keypoint estimation involves correcting mispredicted keypoints via user interaction.
 The task aims to reduce the number of required user modifications compared to manual revision by automatically revising all mispredictions given user feedback to a few. In our work, manual revision indicates fully-manually revised results by a user without the assistance of an interactive model.

\textbf{Keypoint coordinate encoding.}
Let $C$, $W$, and $H$ denote the number of channels, width, and height of an input image $\mathcal{I}\in \mathds{R}^{C \times W \times H}$.
Given $K$ keypoints to estimate, the groundtruth x-y coordinates of the keypoints are encoded to a Gaussian-smoothed heatmap $\mathcal{H} \in \mathds{R}^{K \times W \times H}$, as used in prior work~\cite{K11,hrnetv2}.

\textbf{User-interaction encoding.} 
User interaction is fed to the model to obtain the revised results.
It is encoded into a $K$-channel heatmap $\mathcal{U} = \{\mathcal{U}_1, \mathcal{U}_2, ..., \mathcal{U}_K\}$, where $\mathcal{U}_n \in \mathds{R}^{1 \times W\times H}$ is revision information for the $n$-th keypoint. Given $l$ user modifications to a subset of keypoints, $\{c_1, c_2, ... c_l\}$, the corresponding channels $\{\mathcal{U}_{c_1}, \mathcal{U}_{c_2}, ..., \mathcal{U}_{c_l}\}$ are {activated} as the Gaussian-smoothed heatmap, whereas the other channels corresponding to the unmodified keypoints are filled with zeros. Formally, a user interaction heatmap for the $n$-th keypoint can be expressed as 
\begin{equation}
  \mathcal{U}_n(i,j) = \begin{cases}
     \exp\Big(\frac{(i-x_n)^2+(j-y_n)^2}{-2\sigma^2}\Big), & \text{if $n\in \{c_1, c_2, ..., c_l\}$}.\\
    0, & \text{otherwise},
  \end{cases}
\end{equation}
where $p_n=(x_n, y_n)$ denotes user-modified coordinates of the $n$-th keypoint.

\textbf{Keypoint coordinate decoding.}
Following Bulat et al.~\cite{subpixel}, we transform the predicted heatmaps into 2D keypoint coordinates by using differentiable local soft-argmax to reduce quantization errors caused by the heatmap.

\textbf{Synthesizing user interaction during training.}
During training, we simulate user interaction by randomly sampling a subset of groundtruth keypoints as user-modified keypoints. 
First, we define a multinomial distribution of the number of user modifications with a range of $[0, K]$. Since our goal is to correct errors with a small number of modifications, we set the probability of obtaining a larger number of modifications to become exponentially smaller.
Once the total number of modifications is determined, the keypoints to be corrected are randomly sampled from a discrete uniform distribution $\text{Unif}(1, K)$. 

\textbf{Network architecture.}
As illustrated in Fig.~\ref{fig:main_network}, we employ a simple convolutional block, called hint fusion layer, to feed additional input to the model, e.g., user clicks, without any architectural changes to the backbone following RITM~\cite{ritm}. The hint fusion layer receives an input image, user interaction, and previous prediction of the model and
returns a tensor having the same shape as the output of the first block of the backbone network.
Also, we adopt an iterative training procedure so that users can repeatedly revise a keypoint prediction result. Meanwhile, the previous prediction of the model is fed to the model in the next step to make it aware of its earlier predictions~\cite{mahadevanitis2018,ritm}. During the inference time, we extend this to selectively provide previous predictions just for user-modified keypoints, not all keypoints, to facilitate differentiating the mispredictions. We use the high-resolution network (HRNet)-W32~\cite{hrnetv2} with object-contextual representations (OCR)~\cite{ocr} as a pre-trained backbone network. 
The model is trained with a binary cross-entropy loss $L_{g}$ between the predicted heatmap $\hat{\mathcal{H}}$ and the Gaussian-smoothed target heatmap ${\mathcal{H}}$. As a final prediction, we post-process the predictions so that a user-modified point stays where the user wants it.

\subsection{Interaction-Guided Gating Network}\label{attention}
We propose an interaction-guided gating network, which effectively propagates user-interaction information throughout the entire spatial area of an image.
Given user interaction $\mathcal{U}$ and the downsampled feature map $\mathbf{F}_h$ from the backbone network, the proposed method generates a channel-wise gating weight $\mathbf{A} \in \mathds{R}^{k_c}$ to recalibrate the intermediate feature map $\mathbf{F}_c \in \mathds{R}^{k_c\times W' \times H'}$ according to user feedback summarized globally from every pixel position. 
Inspired by Hu et al.~\cite{squeezeAndExcitation}, which use global average pooling to aggregate global spatial information, we employ global max-pooling to selectively retrieve significant interaction-aware features from the entire spatial area for each channel. Then, the gating weight $\mathbf{A}$ is generated by subsequent fully connected layers and a sigmoid activation function and utilized to gate the feature map $F_c$ channel-wise.
Finally, the reweighted feature map is decoded into a keypoint heatmap by our prediction head.

\begin{figure*}[t!]
\begin{center}
  \includegraphics[width=1.0\linewidth]{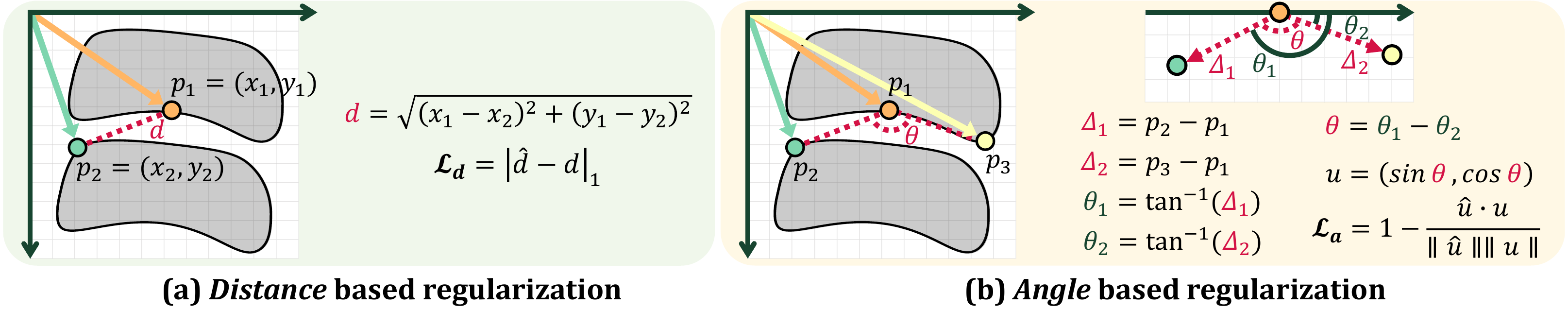}
\end{center}
  \caption{{Illustration of Morphology-aware loss.}} 
\label{fig:morph_overview}
\end{figure*}

\subsection{Morphology-Aware Loss}\label{morph}
The morphology-aware loss aims to explicitly utilize the inter-keypoint relationships as well as the individual information of each keypoint to make the model aware of the morphological associations among keypoints.
Specifically, we exploit two higher-order statistics: the distance between two keypoints and the angle among three keypoints, as shown in Fig~\ref{fig:morph_overview}. 
Given $K$ points, the numbers of all possible combinations of keypoint sets are ${K\choose 2}$ for the distance and ${K\choose 3}$ for the angle.
Instead of using all possible sets, we leverage only the ones that rarely deviate across the dataset, assuming that learning significantly varying relationships can degrade the model performance.
Thus, we select two subsets to apply the morphology-aware loss based on the standard deviation values of the distance and angle. 
The standard deviation values can be obtained as 
\begin{align}
S_d = \sqrt{\E[(d-\E[d])^2]}  \text{\ \ and\ \ } 
S_a = \sqrt{-\ln(\E[u_x]^2+\E[u_y]^2)},
\end{align}
where $d$ and $u=(u_x, u_y)$ indicate the distance and angle vector defined in Fig.~\ref{fig:morph_overview}, respectively. 
Using the standard deviation values as criterion to select the elements, the subsets of keypoint relationships can be obtained as
\begin{equation}
\begin{aligned}
\mathcal{P}_d &= \{(p_m, p_n)  \ | \   S_d\big(d(p_m, p_n)\big) < t_d,\    m \neq n;\  
m, n \in {[1, K]} \} \text{\ \ and}  \\
\mathcal{P}_a &= \{(p_m, p_n, p_l) \  |\   S_a\big(u(p_m, p_n, p_l)\big) < t_a,\    m \neq n \neq l;\   
 m, n, l \in [1,K]\},
\end{aligned}
\end{equation}
where $t_d$ and $t_a$ are threshold values. Here, the circular variance is computed for the angle.
Finally, the morphology-aware loss is defined as $L_m = L_{d} + \lambda_m L_{a}$, where $L_{d}$ is the $L1$ loss for distance applied to the set $\mathcal{P}_d$, and $L_{a}$ is the cosine similarity loss for angle vector applied to the set $\mathcal{P}_a$. Ultimately, when a user revises one keypoint, the keypoints having regular relationships with the revised one will also be updated to preserve the inter-keypoint relationships in our model.

\section{Experiments}

\textbf{Dataset.}
We collect our own dataset, called the \textit{\ourdata} dataset, which contains 6,504 cephalometric X-ray images (677/692/5,135 for training/validation/test) of 4,280 subjects without overlapping patients in each set. 
We use a small training set and large test set in consideration of the medical domain, in which reliable performance validation is critical, and data available for training is scarce~\cite{cvpr_medical_1,cvpr_medical_3,cvpr_medical_2}.
Each image is annotated with 13 keypoints, which include the vertex points of the cervical vertebrae. The keypoints can be used to examine the skeletal growth of a patient by calculating anatomical measurements such as the concavity of the vertebrae~\cite{kim2021prediction,med_1}.
The \textit{\spinewebdata} dataset~\cite{spinewebdataset} contains spinal anterior-posterior X-ray images (352/128/128 for training/validation/test). In each image, 68 points representing the four vertices of 17 vertebrae are annotated. 

\textbf{Evaluation metrics.}
We measure the mean radial error (MRE), which is calculated as $\text{MRE}=\frac{1}{K}\sum_{n=1}^K(||p_n-\hat{p}_n||_2)$.
Also, we borrow the evaluation protocol of interactive segmentation tasks~\cite{firstclick,fbrs}, number of clicks (NoC), to evaluate the performance of the proposed interactive keypoint estimation approach.
In our work, NoC measures the average number of clicks required to achieve a target MRE. 
We set ranges of target MRE as $[0, 10]$ and $[0, 60]$ for \ourdata and \spinewebdata, respectively. We report selected results in Table~\ref{table:sota}, and the results for different target MRE values are provided in Fig.~6 of the supplementary material.
When measuring NoC, we limit the maximum number of user modifications to a small value because we aim to revise all mispredictions with as few clicks as possible. Since \ourdata and \spinewebdata include 13 and 68 target keypoints, we set the limits to five and ten clicks, respectively.
Finally, $\text{NoC}_{\alpha}@\beta$ denotes the average number of clicks to achieve the target MRE of $\beta$ pixels when a prediction of an image can be maximally modified by up to $\alpha$ clicks. Similarly, $\text{FR}_{\alpha}@\beta$ counts the average rate of images that fail to achieve the target MRE of $\beta$ pixels when the model is given a maximum of $\alpha$ clicks. 
Under the assumption that a doctor would correct clearly wrong keypoints first, the keypoint having the highest error for each prediction is selected to revise in all experiments.

\textbf{Experimental Details.}
We resize the images to $512 \times 512$ and $512 \times 256$ for the Cephalometric X-ray and \spinewebdata datasets, respectively. All models used in our experiments are implemented in PyTorch~\cite{pytorch} and experimented on a single GeForce RTX 3090 in the training and evaluation phases.
The proposed model has $32.9M$ trainable parameters. 
The model is trained using the Adam optimizer~\cite{kingma2014adam} with a learning rate of $0.001$ and a batch size of 4.
The training is early-stopped when mean radial errors do not decrease for 50 epochs on the validation dataset.
More details on hyperparameter settings of the proposed model are available at our \href{https://github.com/seharanul17/interactive_keypoint_estimation}{\textbf{GitHub repository}}.

\textbf{Compared baselines.}
We assess the proposed interactive keypoint estimation model by comparing it with state-of-the-art interactive segmentation approaches, which share similar concepts in the interactive system: BRS~\cite{brs}, f-BRS~\cite{fbrs}, and RITM~\cite{ritm}. Along with the ablation study, this is an essential evaluation step in our method, given the lack of research on interactive keypoint estimation.
The baselines encode user feedback as distance maps~\cite{brs,fbrs} or hard masks~\cite{ritm} tailored to interactive segmentation, so we replace them with our $K$-channel user interaction heatmap in the experiments. 
Additionally, we modify the baselines to output a $K$-channel heatmap and to minimize the binary cross-entropy loss between the predicted and groundtruth heatmap.
For BRS and f-BRS, we observe that their backpropagation refinement schemes degrade model performance in our task; we report their results without the schemes.

\begin{table}[!t]
\caption{{Comparison with baselines on \ourdata and \spinewebdata.}}
\label{table:sota}
\begin{center}
\resizebox{0.73\textwidth}{!}{%
\begin{tabular}{l||ccccc|ccccc}
\toprule
\multicolumn{1}{c}{\multirow{3.3}{*}{Method}} & 
\multicolumn{5}{c}{\ourdata} & 
\multicolumn{5}{c}{\spinewebdata} \\
\cmidrule(l{2pt}r{2pt}){2-6} \cmidrule(l{2pt}r{2pt}){7-11}  
\multicolumn{1}{c}{}&
\makecell{$\text{FR}_5$\\@3} & 
\makecell{$\text{NoC}_5$\\@3} &
\makecell{$\text{NoC}_5$\\@4} & 
\makecell{$\text{NoC}_5$\\@5} &
\makecell{$\text{NoC}_5$\\@6} & 
\makecell{$\text{FR}_{10}$\\@20} &
\makecell{$\text{NoC}_{10}$\\@20} &
\makecell{$\text{NoC}_{10}$\\@30} &
\makecell{$\text{NoC}_{10}$\\@40} &
\makecell{$\text{NoC}_{10}$\\@50} \\
\midrule\midrule
\multicolumn{1}{l||}{$\text{BRS}$~\cite{brs}} &9.72 &  3.18 & 1.65 & 0.78 & 0.36 & 11.72 & 3.02 & 2.41 & 1.91 & 1.59\\
\multicolumn{1}{l||}{$\text{f-BRS}$~\cite{fbrs}} & 9.80 & 2.99 & 1.34 & 0.52 & 0.22 & 52.34 & 7.36 & 5.61 & 4.55 & 3.82\\
\multicolumn{1}{l||}{RITM~\cite{ritm}} & 10.69 & 3.12 & 1.48 & 0.60 & 0.25 & 13.28 & 3.56 & 2.73 & 2.14 & 1.56\\
\midrule  
\multicolumn{1}{c||}{\textbf{\ourmethod}}& \textbf{4.48} & \textbf{2.32} & \textbf{0.86} & \textbf{0.31} & \textbf{0.13}  & \textbf{6.25} & \textbf{2.46} & \textbf{1.88} & \textbf{1.41} & \textbf{1.19}\\
\bottomrule
\end{tabular}}
\end{center}
\end{table}

\begin{figure}[!t]
\begin{center}
\includegraphics[width=1.0\linewidth]{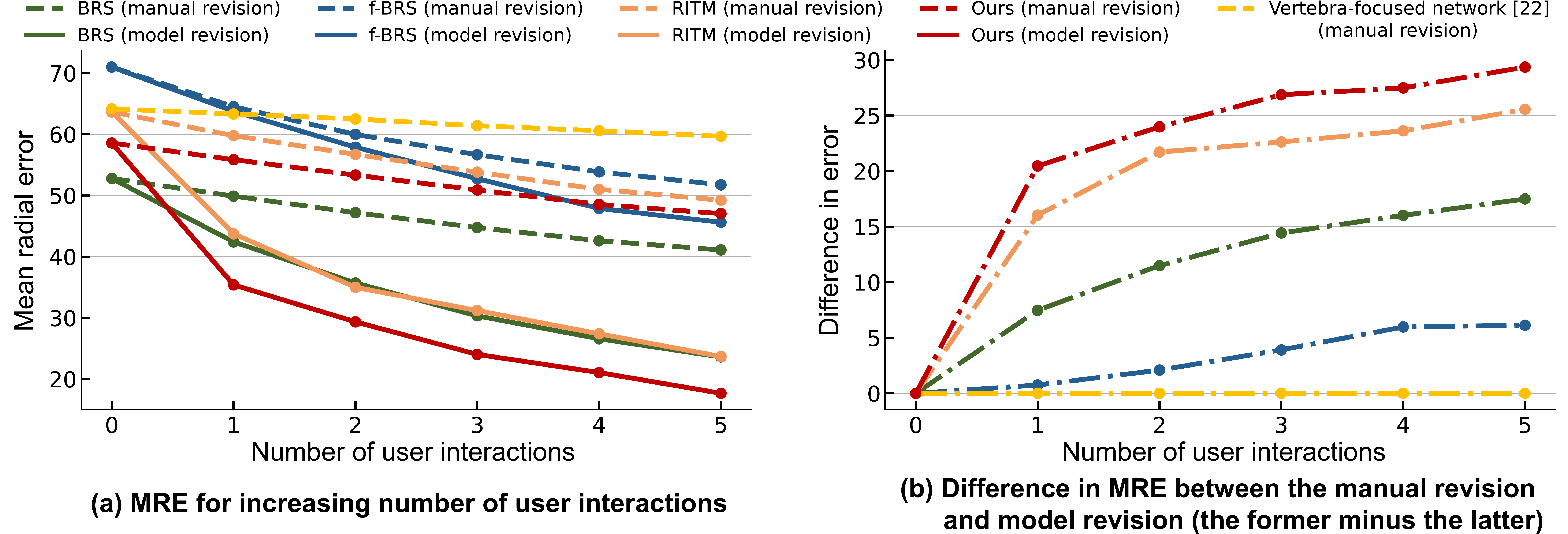}
\end{center}
    \caption{{Comparison with manual revision on \spinewebdata.}}
\label{fig:error_histogram}
\end{figure}

\begin{figure}[t!]
\begin{center}
\includegraphics[width=1.0\linewidth]{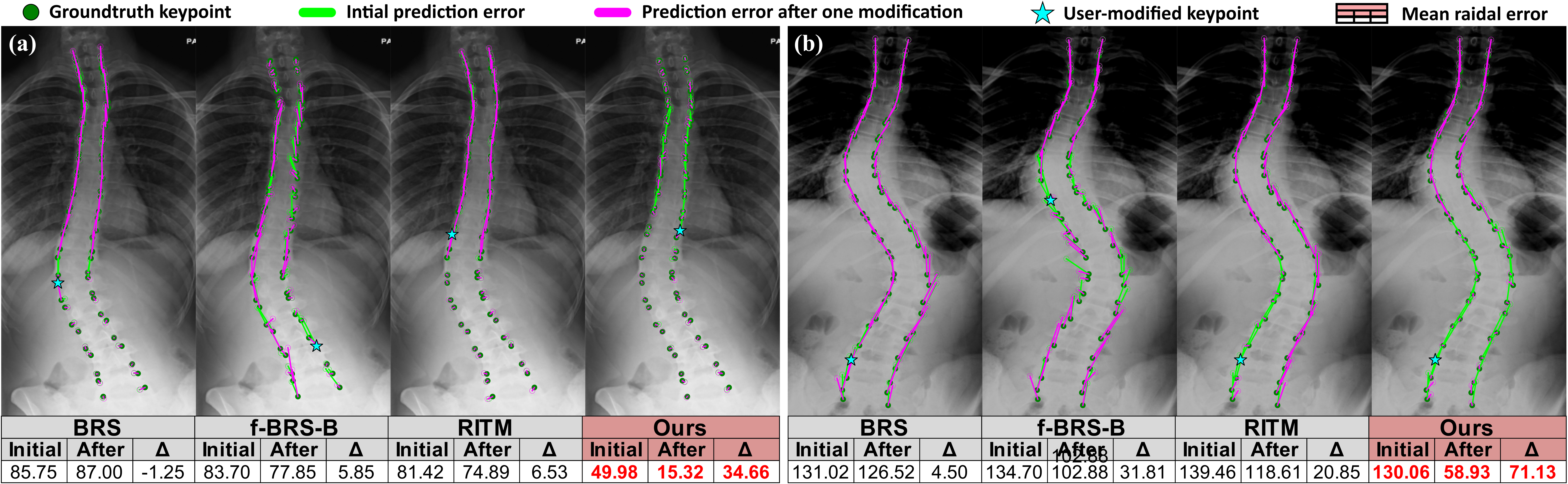}
\end{center}
  \caption{{Qualitative interactive keypoint estimation results on \spinewebdata.} 
  To visualize the prediction error, we draw a line between the predicted keypoints and the corresponding groundtruth keypoints; the shorter, the better.
  Given user feedback, \textit{the area where the green lines are dominant} is where errors are significantly reduced compared to initial predictions.
  \texttt{Initial}, initial prediction error; \texttt{After}, prediction error after one user modification; \texttt{$\Delta$}, \texttt{Initial minus After}.
  }
\label{fig:quality}
\end{figure}

\begin{table}[t!]
\footnotesize
\caption{{Ablation study of our method on \ourdata}.
Low variance, in which the morphology-aware loss is applied to the \textit{top} 15 items having the \textit{lowest variance} among all possible items; high variance, in which the morphology-aware loss is applied to the \textit{top} 15 items having the \textit{highest variance} among all; adjacent points, in which all keypoint sets comprise an \textit{edge or internal angle of vertebrae}.}
\label{table:lonperformances}
\begin{center}\resizebox{0.8\textwidth}{!}{%
\begin{tabular}{ccc|cc||cccc}
\toprule
\multicolumn{3}{c}{\oursemethod} &
\multicolumn{2}{c}{\multirow{1}{*}{\makecell{Morphology-aware loss}}}&
\multicolumn{4}{c}{Performance} \\
\cmidrule(lr ){1-3}
\cmidrule(lr ){4-5}
\cmidrule(lr ){6-9}
\multicolumn{1}{c}{} &
\multicolumn{1}{c}{\makecell{Pooling\\method}} & 
\multicolumn{1}{c}{\makecell{Activation\\function}} & 
\multicolumn{1}{c}{} &
\multicolumn{1}{c}{\makecell{Criterion for \\ $\mathcal{P}_d$ and $\mathcal{P}_a$}} &
\multicolumn{1}{c}{\makecell{$\text{FR}_{5}$\\$@3$}}&
\multicolumn{1}{c}{\makecell{$\text{NoC}_{5}$\\$@3$}}&
\multicolumn{1}{c}{\makecell{$\text{NoC}_{5}$\\$@4$}}&
\multicolumn{1}{c}{\makecell{$\text{NoC}_{5}$\\$@5$}}\\
\midrule \midrule
 \xmark &- & - & \xmark & -  &  10.09 & 3.09 & 1.46 & 0.60\\ 
\cmark& max &sigmoid& \xmark &- & 5.88 & 2.51 & 0.98 & 0.37 \\ 
\textbf{\cmark}& \textbf{max} &\textbf{sigmoid}  &\textbf{\cmark} &\textbf{low variance}  & \textbf{4.48} & \textbf{2.32} & \textbf{0.86} & \textbf{0.31} \\ 
\cmidrule(l{1.5pt}r{1.5pt}){1-9} 
  \xmark &- & - & \cmark & low variance  & 6.33  & 2.56 & 1.03 & 0.38\\ 
\cmark& average &sigmoid  &\cmark&low variance & 5.39 & 2.53 & 1.00 & 0.37 \\ 
\cmark& max  &softmax& \cmark & low variance & 5.71 & 2.54 &0.96 & 0.34 \\
\cmidrule(l{1.5pt}r{1.5pt}){1-9} 
\cmark & max &sigmoid  &\cmark &high variance & 6.23 & 2.62 & 1.04 & 0.38\\
 \cmark & max &sigmoid  &\cmark &adjacent points & 4.79 & 2.44 & 0.93 & 0.33 \\
\bottomrule
\end{tabular}}
\end{center}
\label{table:abalation}
\end{table}

\textbf{Quantitative comparison with baselines.} 
Table~\ref{table:sota} shows that our method consistently outperforms the baselines in terms of the number of clicks (NoC) and failure rate (FR) by a significant margin.
 For example, our method reduces the failure rate by about half compared to the baselines. 
Moreover, the mean number of clicks of our method is consistently lower than that of the baselines for both datasets.
In Fig.~\ref{fig:error_histogram}, when the proposed interactive keypoint estimation approach (\textit{model revision}) is applied, the prediction error is significantly reduced compared to the fully manually revised result (\textit{manual revision}). 
The superiority of the proposed framework is also revealed by the comparison with a baseline model without the interactive framework, the Vertebra-focused network~\cite{vertebra-focused}.
The error reduction rate of the network is notably slow when its predictions are manually revised.
Also, among all models, our model achieves the largest improvement compared to manual revision and consistently outperforms RITM by a large margin, decreasing the error by five pixels on average. 
Altogether, our approach successfully propagates user revision information to unrevised but inaccurate keypoints, demonstrating the effectiveness of the proposed interactive framework.

\textbf{Qualitative comparison with baselines.}
Fig.~\ref{fig:quality} illustrates the prediction results on \spinewebdata with their mean radial errors before and after user revision.
Our proposed model remarkably reduces the initial error in a much wider area than the baselines, successfully propagating user feedback to distant mispredictions.
%

\textbf{Ablation study.}
Table~\ref{table:abalation} validates each component of our method. The results demonstrate that our model outperforms the ablated versions.
We ablate the distance and angle sets criterion to apply the morphology-aware loss, $\mathcal{P}_d$ and $\mathcal{P}_a$. 
As a result, learning highly varying inter-keypoint relationships degrades the model performance contrary to learning rarely-varying relationships.
Results for different threshold values are provided in Table~4 of the supplementary material.

\section{Conclusion}
This paper focuses on improving the efficiency and usability of the keypoint annotation to speed up the overall process while improving diagnostic accuracy.
To this end, we introduce a novel interactive keypoint estimation network, incorporating the interaction-guided attention network and the morphology-aware loss to revise inaccurate keypoints with a small number of user modifications. We demonstrate the effectiveness of the proposed approach through extensive experiments and analysis of two medical datasets.
This work assumes that a user will always provide a correct modification to the model and revise a clearly wrong keypoint first. Thus, future work can address noisy interactive inputs by real users or find the most effective keypoint to correct all other keypoints and recommend users to revise it first.
To the best of our knowledge, this is the first work to propose an interactive keypoint estimation framework, and it will be helpful to researchers working on human-in-the-loop keypoint annotation.

\bigskip

\textbf{Acknowledgements.}
This work was supported by the Institute of Information \& communications Technology Planning \& Evaluation (IITP) grant funded by the Korean government(MSIT) (No. 2019-0-00075, Artificial Intelligence Graduate School Program(KAIST)), the National Research Foundation of Korea (NRF) grant funded by the Korean government (MSIT) (No. NRF-2019R1A2C4070420), the National Supercomputing Center with supercomputing resources including technical support (KSC-2022-CRE-0119), and the Korea Medical Device Development Fund grant funded by the Korea government (the Ministry of Science and ICT, the Ministry of Trade, Industry and Energy, the Ministry of Health \& Welfare, the Ministry of Food and Drug Safety) (Project Number: 1711139098, RS-2021-KD000009).


\bibliographystyle{splncs04}
\bibliography{ms}






%
\title{Supplementary Material}
\titlerunning{Morphology-Aware Interactive Keypoint Estimation}
\author{} 
\authorrunning{Morphology-Aware Interactive Keypoint Estimation} 
\institute{}

\maketitle              

\setcounter{table}{2}
\setcounter{figure}{5}

\begin{table}[h]
\caption{Summary of notations used in our paper.}
\label{table:summary}
\begin{center}
\resizebox{0.87\textwidth}{!}{%
\begin{tabular}{c||c|l}
\toprule
Notation&
{Dimension}&
{Description}\\
\midrule\midrule
$C$, $W$, $H$ & $\mathds{R}^{1}$&  
channel dimension, width, height of an input image \\
 $K$ & $\mathds{R}^1$&  
 total number of keypoints \\
 $\mathcal{I} $ & $\mathds{R}^{C\times W \times H}$ &
 input image\\
$\mathcal{H}, \hat{\mathcal{H}} $& $\mathds{R}^{K\times W \times H}$& 
groundtruth, predicted keypoint heatmaps  \\
 $\mathcal{U}$ & $\mathds{R}^{K\times W \times H}$ &
 user-interaction heatmap\\
 $\sigma$ &  $\mathds{R}^{1}$&
 standard deviation of the Gaussian-smoothed heatmap $\mathcal{U}$ \\
$p_n=(x_n,y_n)$, $\hat{p}_n$&  $\mathds{R}^{2}$ &
groundtruth, predicted coordinates of the $n$-th keypoint \\
\midrule
\multicolumn{3}{c}{Interaction-guided gating network}\\
\midrule
 $k_{h}$, $k_c$ & $\mathds{R}^{1}$ & channel dimensions of intermediate feature maps\\
 $W', H'$&$\mathds{R}^{1}$&  
width, height of intermediate feature map\\
 $\textbf{F}_h$& $\mathds{R}^{k_{{h}} \times W' \times H' } $&
 intermediate feature map of the backbone network \\
 $\textbf{F}_c$& $\mathds{R}^{k_{c} \times W' \times H' }$ &
 intermediate feature map to reweight by the proposed method\\
 $\textbf{A}$ & $\mathds{R}^{k_c}$&gating weight  \\
\midrule
\multicolumn{3}{c}{Morphology-aware loss} \\
\midrule
$ d, \hat{d} $&$ \mathds{R}^{1}$ &groundtruth, predicted distances between two keypoints\\ 
$ u=(u_x, u_y),\hat{u} $&$ \mathds{R}^{2}$& groundtruth, predicted angle vectors between three keypoints\\ 
$S_d, S_a$&$\mathds{R}^{1}$&standard deviation values of distance $d$, angle vector $u$\\
$\mathcal{P}_d, \mathcal{P}_a$&-& distance, angle sets to apply the proposed loss\\
${t}_d, {t}_a$& $\mathds{R}^{1}$ & threshold values for $S_d, S_a$ to determine $\mathcal{P}_d, \mathcal{P}_a$\\
$\lambda_m$, $\lambda_t$ & $\mathds{R}^{1}$& loss coefficients that range from zero to one. \\
$L_g$ & $\mathds{R}^{1}$& binary cross-entropy loss for keypoint heatmap\\
$L_d$ & $\mathds{R}^{1}$&  L1 loss for distance\\
$L_a$ &$\mathds{R}^{1}$ &  cosine similarity loss for angle vector \\
$L_m=L_d+\lambda_m L_a$ & $\mathds{R}^{1}$ & morphology-aware loss \\
$L_t=L_g+\lambda_t L_m$ &$\mathds{R}^{1}$ &  total loss\\
\bottomrule
\end{tabular}}
\end{center}
\end{table}

\begin{figure}[h!]
\begin{center}
\includegraphics[width=0.93\linewidth]{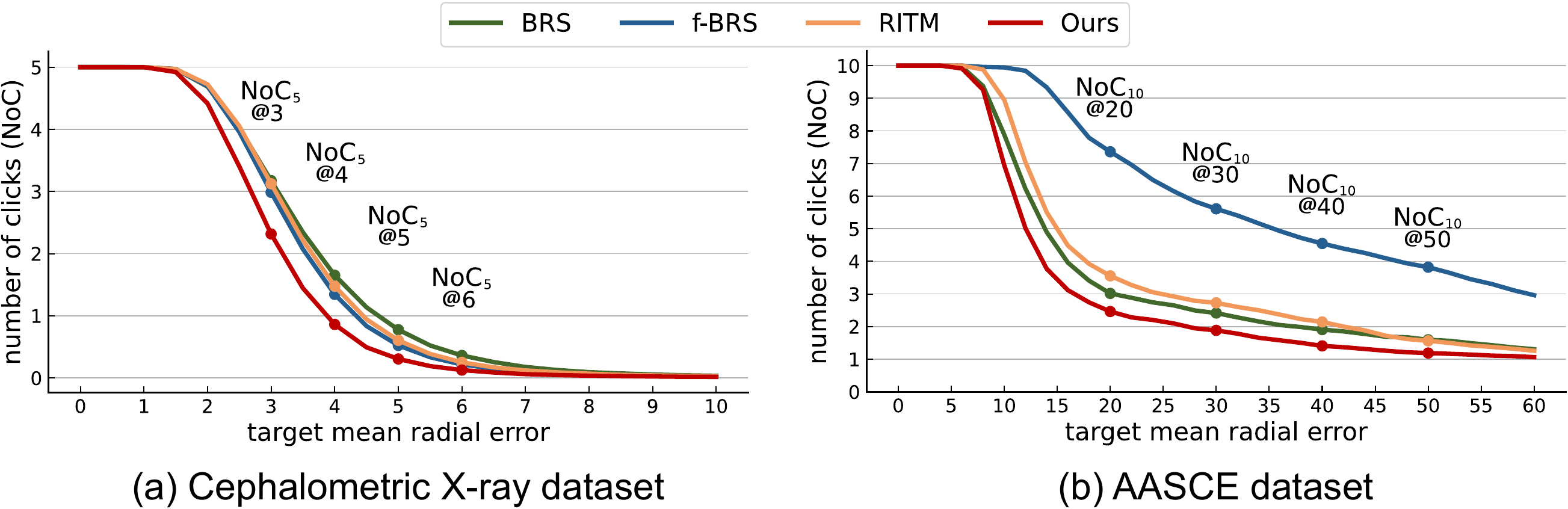}
\end{center}
\caption{Number of user interactions in comparison with baseline models. 
}
\label{suppe_fig:overview}
\end{figure}

\begin{table}[t!]
\footnotesize
\caption{Sensitivity analysis and ablation study of morphology-aware loss on the \spinewebdata dataset.
The failure rate and the number of clicks are measured. 
We prolong the size of the distance and angle sets to apply the proposed loss from 40 to 120 to see its impact on the interactive keypoint estimation performance.
We also compare the proposed loss with its ablated versions: w/o Morph loss, the model trained without the proposed loss $L_m$; r/w Coord loss, the model in which the proposed loss $L_m$ is replaced with an L1 loss between the groundtruth 2D keypoint coordinates and the coordinates extracted from the predicted heatmaps.
}
\label{table:hypsensitivity}
\begin{center}\resizebox{0.85\textwidth}{!}{%
\begin{tabular}{c|ccc|c||ccccc}
\toprule
\multicolumn{1}{c}{\multirow{4.5}{*}{Method}}&
\multicolumn{3}{c}{\multirow{2}{*}{\makecell{Morphology-\\aware loss}}}&
\multicolumn{1}{c}{\multirow{4.5}{*}{\makecell{Coordinate\\ regression \\loss}}} &
\multicolumn{5}{c}{\multirow{2.2}{*}{Performance}} \\
\multicolumn{1}{c}{} &
\multicolumn{1}{c}{} & 
\multicolumn{1}{c}{} &
\multicolumn{1}{c}{} & 
\multicolumn{1}{c}{}\\
\cmidrule(lr ){2-4}
\cmidrule(lr ){6-10}
\multicolumn{1}{c}{} & 
\multicolumn{1}{c}{} &
\multicolumn{1}{c}{\makecell{no.\\$< t_d$}} &
\multicolumn{1}{c}{\makecell{no.\\$< t_a$}} &
\multicolumn{1}{c}{} &
\multicolumn{1}{c}{\makecell{$\text{FR}_{10}$\\$@20$}}&
\multicolumn{1}{c}{\makecell{$\text{NoC}_{10}$\\$@20$}}&
\multicolumn{1}{c}{\makecell{$\text{NoC}_{10}$\\$@30$}}&
\multicolumn{1}{c}{\makecell{$\text{NoC}_{10}$\\$@40$}}&
\multicolumn{1}{c}{\makecell{$\text{NoC}_{10}$\\$@50$}}\\
\midrule \midrule
{\multirow{9}{*}{\textbf{Ours}}} & 
\cmark & 40 & 40 & \xmark  & 9.38 &2.60 &2.02 & 1.56 &1.34 \\
&\cmark & 50 & 50 & \xmark  & 8.59 &2.18 &1.69 & 1.34 &1.15 \\
&\cmark & 60 & 60 & \xmark  & 7.81 &2.22 &\textbf{1.59} & \textbf{1.23} &\textbf{0.99} \\
&\cmark & 70 & 70 & \xmark  & \textbf{6.25} &2.46 &1.88 & 1.41 &1.19 \\
&\cmark & 80 & 80 & \xmark  & 7.03 &2.88 &2.18 & 1.62 &1.21 \\
&\cmark & 90 & 90 & \xmark  & 7.81 &2.20 &1.61 & 1.26 &1.01 \\
&\cmark & 100 & 100 & \xmark  &8.59 &2.35 &1.90 & 1.56&1.30 \\
&\cmark & 110 & 110 & \xmark  & \textbf{6.25} &\textbf{2.08} &\textbf{1.59} & 1.24 &1.06 \\
&\cmark & 120 & 120 & \xmark  & 8.59 &2.96 &2.27 & 1.63 &1.29 \\
\midrule
w/o Morph loss&\xmark &- & - & \xmark  & 12.50 &3.68 &2.62 &1.94 &1.54 \\
r/w Coord loss &\xmark &- & -& \cmark  & 9.38 &3.31 &2.41 & 1.76 &1.43 \\
\bottomrule
\end{tabular}}
\end{center}
\label{table:ablation_supple}
\end{table}

\begin{figure}[t]
\begin{center}
\includegraphics[width=1.0\linewidth]{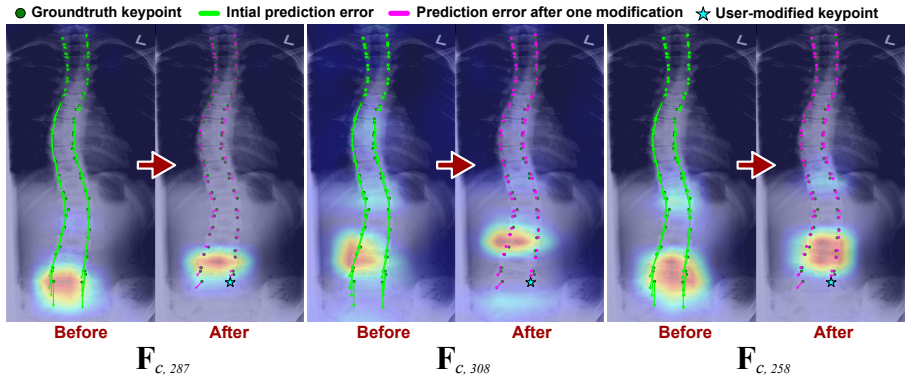}
\end{center}
\caption{Analysis of how user interaction changes each gated feature map in proposed approach.
In this example, the mean radial error is reduced from 90.02 to 10.24 when user interaction is provided to the model. We illustrate the three feature maps having the largest changes in gated feature maps: the 287th, 308th, and 258th feature maps, in descending order.
The model prediction results and corresponding errors (the shorter, the better) are provided in each image as well.
}
\label{suppe_fig:supple_overview}
\end{figure}


\end{document}